\documentclass{article}
\usepackage{style}

\usepackage{times}
\usepackage{soul}
\usepackage{url}
\usepackage[hidelinks]{hyperref}
\usepackage[utf8]{inputenc}
\usepackage[small]{caption}
\usepackage{graphicx}
\usepackage{amsmath}
\usepackage{amsthm}
\usepackage{booktabs}
\usepackage{multirow}
\usepackage{algorithm}
\usepackage{algorithmic}
\usepackage[switch]{lineno}
\usepackage{amsfonts} % for \mathbb
\newcommand{\TODO}[1]{\textbf{TBD}} % placeholder marker

\usepackage{makecell}
\usepackage{tabularx}
\usepackage{threeparttable}
\title{AC$^2$-VLA: Action-Context-Aware Adaptive Computation in Vision-Language-Action Models for Efficient Robotic Manipulation}

\author{
Wenda Yu$^1$
\and
Tianshi Wang$^1$\thanks{Corresponding author.}\and
Fengling Li$^2$\and
Jingjing Li$^3$\and
Lei Zhu$^1$\\
\affiliations
$^1$Tongji University\\
$^2$University of Technology Sydney\\
$^3$University of Electronic Science and Technology of China\\
\emails
yu\_wenda@126.com,
tswang0116@163.com,
fenglingli2023@gmail.com,\\
lijin117@yeah.net,
leizhu0608@gmail.com
}

\begin{document}

\maketitle

\begin{abstract}
Vision-Language-Action (VLA) models have demonstrated strong performance in robotic manipulation, yet their closed-loop deployment is hindered by the high latency and compute cost of repeatedly running large vision-language backbones at every timestep. We observe that VLA inference exhibits structured redundancies across temporal, spatial, and depth dimensions, and that most existing efficiency methods ignore action context, despite its central role in embodied tasks. To address this gap, we propose Action-Context-aware Adaptive Computation for VLA models (AC$^2$-VLA), a unified framework that conditions computation on current visual observations, language instructions, and previous action states. Based on this action-centric context, AC$^2$-VLA adaptively performs cognition reuse across timesteps, token pruning, and selective execution of model components within a unified mechanism. To train the adaptive policy, we introduce an action-guided self-distillation scheme that preserves the behavior of the dense VLA policy while enabling structured sparsification that transfers across tasks and settings. Extensive experiments on robotic manipulation benchmarks show that AC$^2$-VLA achieves up to a 1.79$\times$ speedup while reducing FLOPs to 29.4\% of the dense baseline, with comparable task success. Source codes can be found at \url{https://github.com/SunnyYWD/AC-2-VLA}.
\end{abstract}

\section{Introduction}
\label{sec:intro}

\begin{figure}[t]
    \centering
    \includegraphics[width=\columnwidth]{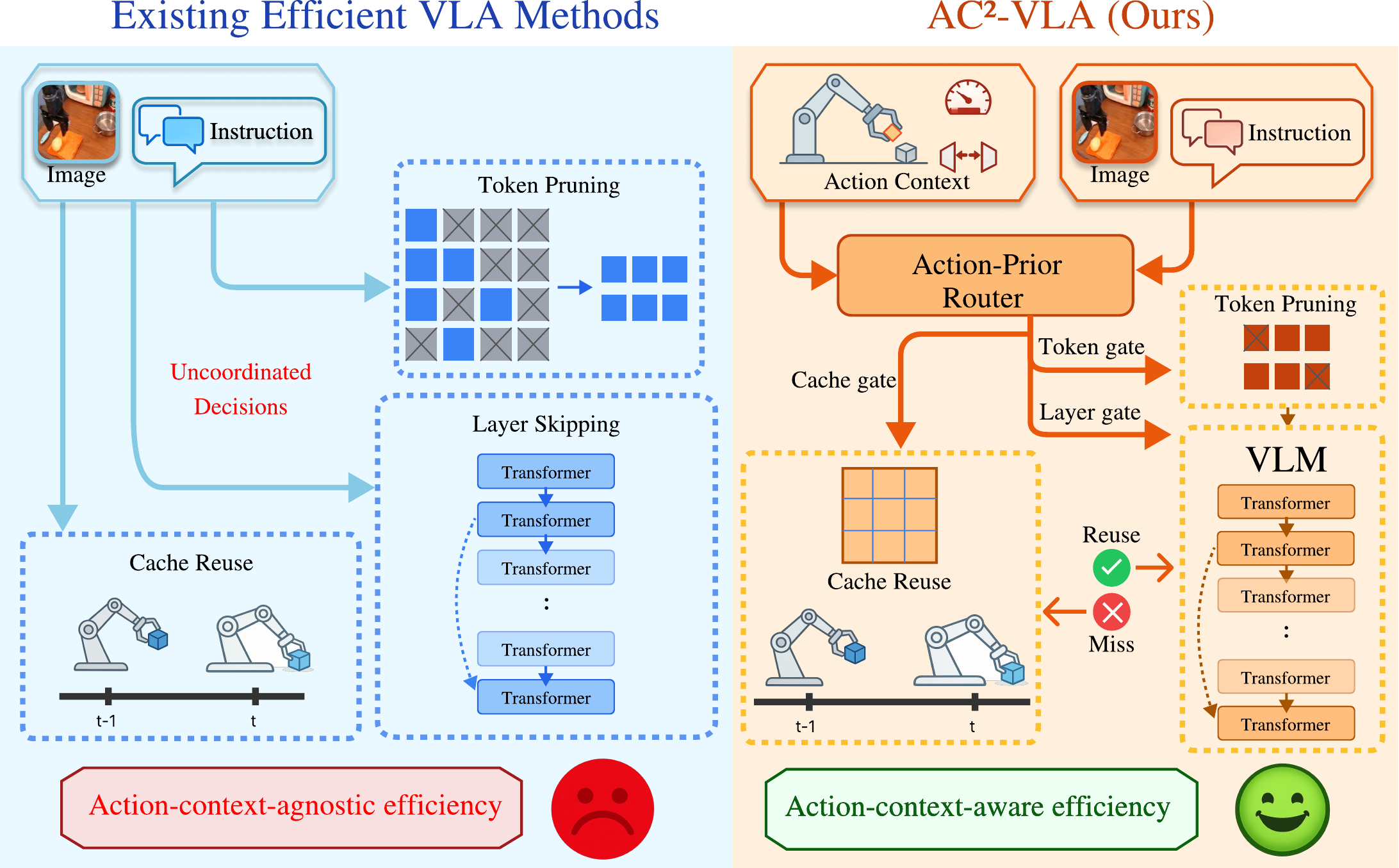}
    \caption{Comparison of efficient VLA computation strategies. Existing methods typically apply cache reuse, token pruning, or layer skipping based on visual or heuristic cues in an uncoordinated manner, resulting in action-context-agnostic efficiency. In contrast, AC$^2$-VLA leverages action context to jointly gate cache reuse, token pruning, and layer skipping for action-context-aware efficiency.}
    \label{fig:intro_overview}
\end{figure}

Recent progress in vision-language foundation models and large-scale robot datasets such as Open X-Embodiment \cite{OpenX} have accelerated the development of generalist Vision-Language-Action (VLA) models. In closed-loop embodied tasks, these models must deliver low-latency decisions with stable closed-loop control over long-horizon tasks. Representative methods such as RT-2 \cite{RT2} and OpenVLA \cite{OpenVLA} demonstrate that large multimodal backbones can follow language instructions and generalize to diverse tasks. More recent policies such as CogACT \cite{CogACT} further improve control by generating expressive action trajectories via diffusion-based modeling. However, deploying these models remains challenging because inference repeatedly executes a computationally expensive vision-language backbone at every control step, resulting in high latency and compute cost, reducing control frequency, and compromising real-time responsiveness in dynamic environments.

To mitigate these deployment challenges, recent work has explored various efficiency mechanisms for VLA models. \textit{Static compression methods} such as pruning \cite{EfficientVLA} and quantization \cite{SQAPVLA} reduce model size but cannot adapt to changing task complexity. \textit{Dynamic computation techniques}, including token pruning \cite{FlashVLA} and layer skipping \cite{MoLeVLA}, adjust compute online, while caching approaches such as VLA-Cache \cite{VLACache} exploit temporal redundancy by reusing features across adjacent timesteps. Despite these advances, most existing methods make compute allocation decisions primarily based on visual cues, which can be suboptimal for robot manipulation. In embodied tasks, visual complexity does not necessarily correlate with control difficulty: visually simple scenes may require full-capacity reasoning for precise interactions, while visually complex transit phases may allow more aggressive pruning. 

Based on this insight, we present Action-Context-aware Adaptive Computation for VLA models (AC$^2$-VLA), as illustrated in Fig.~\ref{fig:intro_overview}. AC$^2$-VLA dynamically allocates computation along the temporal, spatial, and depth dimensions, guided by the action-centric context that is directly relevant to embodied tasks. Specifically, we introduce a lightweight action-prior router that conditions on the previous action state together with multimodal embeddings, and predicts a unified sparsity strategy for the current timestep. The router orchestrates three complementary mechanisms: \textit{(i) cognition caching}, which reuses backbone features across adjacent timesteps when the action context suggests stable state transitions; \textit{(ii) action-context-aware token pruning}, which removes visual tokens that are irrelevant to the current manipulation stage; and \textit{(iii) conditional layer skipping}, which bypasses redundant transformer blocks when the action context indicates lower reasoning demand. To preserve the robustness of the original dense policy under structured sparsification, we train the router using an action-guided self-distillation scheme that encourages consistent action predictions while enabling adaptive computation. Results show that AC$^2$-VLA significantly reduces inference cost while maintaining strong manipulation success rates, highlighting the effectiveness of action-guided adaptive computation for efficient closed-loop control. In summary, our contributions are as follows:
\begin{itemize}
    \item We identify that computation redundancy in VLA models aligns more with action context than visual cues, and propose action-context-aware adaptive computation for efficient robotic manipulation.
    \item We propose AC$^2$-VLA with an action-prior router that adaptively coordinates cognition caching, token pruning, and layer skipping, supported by action-guided self-distillation for robust sparsification.
    \item Experiments on robotic manipulation benchmarks show substantial latency and FLOPs reductions with minimal performance degradation, and confirm consistent gains over baselines with extensive ablations.
\end{itemize}

\section{Related Work}
\label{sec:related_work}

\subsection{Vision-Language-Action Models}
The rapid progress of VLMs, together with large-scale robot data collections such as Open X-Embodiment \cite{OpenX}, has accelerated the emergence of generalist embodied policies that unify perception, instruction following, and action prediction. Among modern VLA systems, the RT series, such as RT-1 and RT-2, demonstrates that token-based autoregressive decoding can scale to broad task sets via large multimodal backbones \cite{RT1,RT2}. Building on similar foundations, OpenVLA \cite{OpenVLA} further systematizes training and evaluation for vision-language-action modeling at scale. In parallel, diffusion- and flow-based policies have become a compelling alternative for continuous control, where action generation is formulated as conditional denoising or flow matching and sampled as coherent trajectories, such as CogACT \cite{CogACT} and the $\pi_0/\pi_{0.5}$ line \cite{pi0,pi05}. Despite their strong generalization and stable closed-loop behaviors, these models share a common deployment bottleneck: regardless of paradigm, VLA inference is dominated by repeatedly executing a large multimodal backbone, leading to high latency in real-time control.

\subsection{Efficient VLA Strategies}
Prior work on efficient VLA strategies broadly falls into four categories: lightweight model design, dynamic routing and conditional execution, compression via pruning and quantization, and temporal reuse through caching or compute switching. Lightweight designs reduce per-step cost by scaling down backbones or streamlining training and inference, such as TinyVLA \cite{TinyVLA}, SmolVLA \cite{SmolVLA}, and FLOWER \cite{FLOWER}. Dynamic routing methods reduce computation by activating only a subset of the model, such as MoLe-VLA \cite{MoLeVLA} and instruction-driven routing with structured sparsification in CogVLA \cite{CogVLA}. Compression-oriented approaches aim to remove redundant tokens and layers or co-design pruning with quantization, such as LightVLA \cite{LightVLA}, FlashVLA \cite{FlashVLA}, and SQAP-VLA \cite{SQAPVLA}. Temporal reuse exploits redundancy across adjacent control steps by reusing cognition or switching computation modes, such as VLA-Cache \cite{VLACache}, SP-VLA \cite{SPVLA}, and VOTE \cite{VOTE}. However, existing methods often target only one redundancy axis or rely on static heuristics without modeling action context, limiting robustness in closed-loop control. In contrast, AC$^2$-VLA uses action context for routing and unifies temporal reuse, spatial sparsification, and depth-wise conditional execution within an action-prior router.

\begin{figure*}[t]
    \centering
    \includegraphics[width=\textwidth]{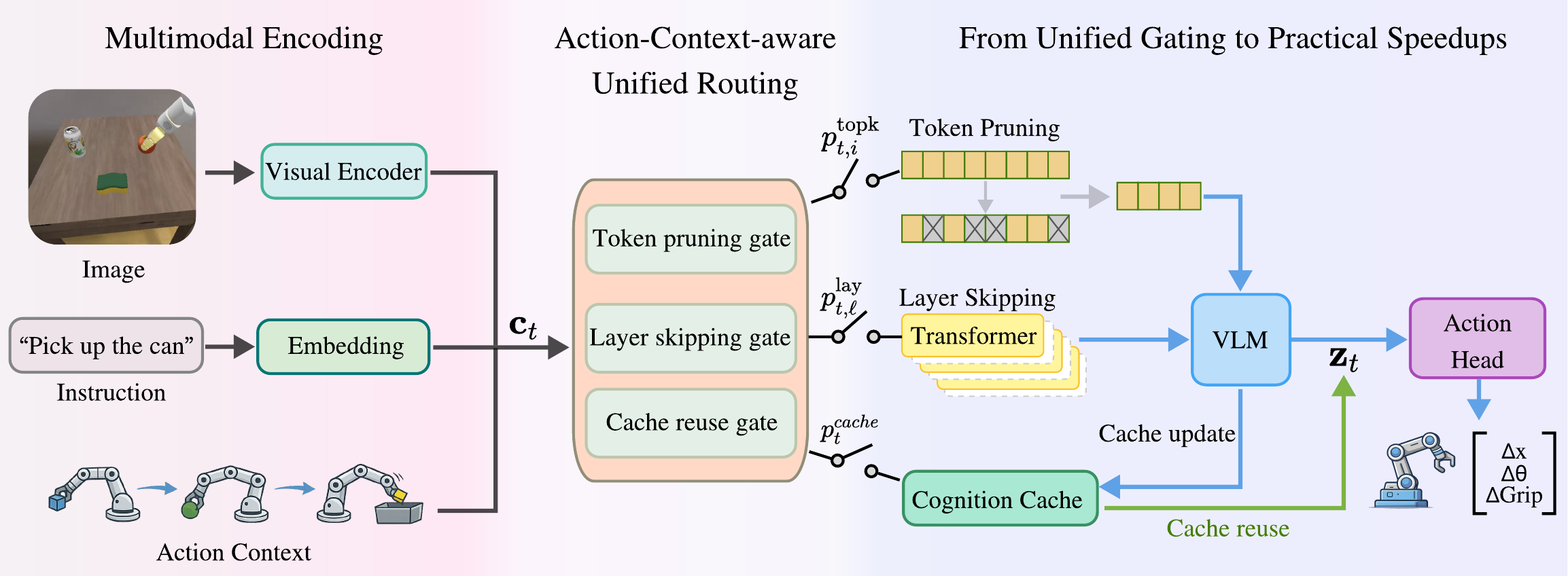}
	    \caption{Overview of the proposed AC$^2$-VLA. At each timestep, the model builds an action-prior condition $\mathbf{c}_t$ from the current observation, instruction, and action context, and uses a unified router to generate token pruning, layer skipping, and cache reuse gates, enabling efficient computation and low-latency control.}
	    \label{fig:ap_framework}
\end{figure*}

\section{Method}
\label{sec:method}

\subsection{Overview}
\label{sec:method_overview}
Given a visual observation $x_t$ and a language instruction $u$, a VLA model predicts an action chunk $\mathbf{a}_{t:t+H}$ with horizon $H$. We consider a generic VLA pipeline that factorizes action generation into a multimodal backbone and an action head:
\begin{equation}
\mathbf{z}_t = f_{\mathrm{VLM}}(x_t,u),
\qquad
\mathbf{a}_{t:t+H}\sim p_{\phi}(\mathbf{a}\mid \mathbf{z}_t).
\label{eq:backbone_factor}
\end{equation}
Here, $p_{\phi}$ can be instantiated as an autoregressive decoder or a diffusion/flow-based trajectory generator, depending on the underlying VLA policy. 

In real-time deployment, inference is bottlenecked by repeatedly executing the VLM backbone at every control step. We observe structured computation redundancies along three complementary axes: 
{\renewcommand{\labelenumi}{(\textit{\roman{enumi})}}
\begin{enumerate}
\setlength{\topsep}{0pt}
\setlength{\itemsep}{0pt}
\item \textit{Temporal redundancy}: backbone representations can be reused across adjacent timesteps;
\item \textit{Spatial redundancy}: only a subset of vision tokens is necessary for action prediction;
\item \textit{Depth redundancy}: executing fewer backbone layers often suffices with minimal performance loss.
\end{enumerate}}

To exploit these redundancies within a unified mechanism, AC$^2$-VLA introduces an action-prior router, as shown in Fig.~\ref{fig:ap_framework}. At each timestep, conditioned on the action-centric context $\mathbf{c}_t$, an action-prior router generates a set of computation gates for temporal reuse, spatial token selection, and depth-wise conditional execution:
\begin{equation}
\begin{aligned}
\mathbf{p}_t=&[p^{\mathrm{cache}}_t; \ \mathbf{p}^{\mathrm{topk}}_t;\ \mathbf{p}^{\mathrm{lay}}_t], \\
p^{\mathrm{cache}}_t \in [0,1], \ \ \ \
&\mathbf{p}^{\mathrm{topk}}_t \in [0,1]^{N_v},\ \ \ \
\mathbf{p}^{\mathrm{lay}}_t \in [0,1]^L,
\end{aligned}
\label{eq:arch_gates}
\end{equation}
where $N_v$ and $L$ denote the numbers of vision tokens and transformer layers, respectively. The cache gate $p^{\mathrm{cache}}_t$ determines whether to reuse cached backbone representations, while $\mathbf{p}^{\mathrm{topk}}_t$ and $\mathbf{p}^{\mathrm{lay}}_t$ control token pruning and conditional layer execution. We train the router via teacher-student distillation with lightweight regularization to preserve dense-policy behavior under structured sparsification.

Overall, AC$^2$-VLA jointly optimizes temporal reuse, token selection, and conditional layer execution, enabling efficient inference for closed-loop robotic manipulation.

\subsection{Action-Context-aware Unified Routing}
\label{sec:method_router}
AC$^2$-VLA is driven by an action-prior condition vector $\mathbf{c}_t$ that explicitly encodes the robot's action context. In closed-loop control, the next action distribution is strongly shaped by the ongoing motion state, making the previous action $\mathbf{a}_{t-1}$ a natural and inexpensive prior for allocating computation. We therefore use $\mathbf{a}_{t-1}$, parameterized consistently with the action head, as the primary routing signal. When no previous action is available at the first step, we set $\mathbf{a}_{t-1}=\mathbf{0}$ and rely on lightweight visual and instruction summaries to generate initial gates.

Let $\mathbf{V}_t\in\mathbb{R}^{N_v\times d_v}$ denote per-token vision features, and we summarize them with a mean-max mixture:
\begin{equation}
\mathbf{s}^{v}_t=
\tfrac{1}{2}\!\left(\mathrm{MeanPool}(\mathbf{V}_t)+\mathrm{MaxPool}(\mathbf{V}_t)\right).
\label{eq:cond_vis}
\end{equation}
For language, we avoid an additional full forward by pooling embedded instruction tokens $\mathbf{E}_t\in\mathbb{R}^{T\times d}$:
\begin{equation}
\mathbf{s}^{u}_t=
\tfrac{1}{2}\!\left(\mathbf{E}_t[\ell_t]+\mathrm{MeanPool}(\mathbf{E}_t)\right),
\label{eq:cond_txt}
\end{equation}
where $\ell_t$ denotes the last valid token index under the attention mask when available, otherwise we use mean pooling.

We embed the action-head step index $\tau_t$ with a sinusoidal encoder $\mathbf{e}(\tau_t)$, which captures the internal generation progress of the action head.
When reuse is enabled, we additionally include a cache-state cue $\mathbf{s}^{c}_t$ encoding the quantized action-delta proxy used for cache keying, together with compact cache statistics and an availability probe.
All inputs are projected to a shared hidden size and fused by an MLP:
\begin{equation}
\begin{aligned}
\mathbf{c}_t = f_{\mathrm{fuse}}(
&\psi_a(\mathbf{a}_{t-1}),\ \psi_v(\mathbf{s}^{v}_t),\ \psi_u(\mathbf{s}^{u}_t),\\
&\psi_\tau(\mathbf{e}(\tau_t)),\ \psi_c(\mathbf{s}^{c}_t)).
\end{aligned}
\label{eq:cond_fuse}
\end{equation}
In implementation, vision tokens and pooled summaries are detached before entering the router to prevent gradients from flowing into heavyweight backbone components through the routing pathway.

Given $\mathbf{c}_t$, the router predicts three gate families: a reuse probability $p^{\mathrm{cache}}_t$, token keep scores $\mathbf{p}^{\mathrm{topk}}_t$, and layer execution gates $\mathbf{p}^{\mathrm{lay}}_t$. We detail each gate in the following.

\vspace{1mm}
\noindent\textbf{Cache reuse gate.} We predict a scalar reuse probability:
\begin{equation}
p^{\mathrm{cache}}_t=
\sigma(\frac{\mathbf{w}^\top \mathbf{c}_t+b}{T_{\mathrm{cache}}}),
\label{eq:cache_gate}
\end{equation}
where $T_{\mathrm{cache}}$ controls gate sharpness.
A high $p^{\mathrm{cache}}_t$ indicates a reuse request, while an actual reuse occurs only when the cache lookup succeeds.

\vspace{1mm}
\noindent\textbf{Token pruning gate.}
For each vision token $\mathbf{v}_{t,i}$, we predict its keep score via action-conditioned matching:
\begin{equation}
p^{\mathrm{topk}}_{t,i}=
\sigma(\langle W_v \mathbf{v}_{t,i},\ W_c \mathbf{c}_t\rangle +\gamma\, g_{t,i}),
\label{eq:tok_score}
\end{equation}
where $g_{t,i}$ is an optional lightweight bias such as a geometric prior derived from the current observation. During inference, tokens are compacted by keeping the top-ranked ones according to $p^{\mathrm{topk}}_{t,i}$.

\vspace{1mm}
\noindent\textbf{Layer skipping gate.}
We predict per-layer execution probabilities:
\begin{equation}
p^{\mathrm{lay}}_{t,\ell}=
\sigma((W_\ell \mathbf{c}_t+\mathbf{b}_\ell)_\ell),
\qquad \ell=1,\dots,L,
\label{eq:lay_gate}
\end{equation}
with bias initialization that favors near-dense execution early in training.
At runtime, transformer blocks with low $p^{\mathrm{lay}}_{t,\ell}$ are conditionally bypassed to reduce depth-wise computation.

\subsection{From Unified Gating to Practical Speedups}
\label{sec:method_sparse}
We next describe how the unified gates translate into practical inference speedups in AC$^2$-VLA, through feature reuse across timesteps, spatial token pruning with compaction, and depth-wise conditional execution. Algorithm~\ref{alg:ap_infer} summarizes the resulting inference-time procedure.

\vspace{1mm}
\noindent\textbf{Cache reuse.}
\label{sec:method_cache}
When the router predicts a high reuse probability $p^{\mathrm{cache}}_t$, we attempt to bypass the expensive multimodal backbone forward by querying a cognition cache. Specifically, we build a compact and robust cache key that captures both motion continuity and visual consistency. We first pool the vision tokens:
\begin{equation}
\bar{\mathbf{v}}_t=\mathrm{MeanPool}(\mathbf{V}_t),
\label{eq:vision_pool}
\end{equation}
and form the cache key as
\begin{equation}
k_t=(\mathrm{Quant}(\lVert\Delta\mathbf{a}_t\rVert),\ \mathrm{Hash}(\bar{\mathbf{v}}_t)),
\label{eq:cache_key}
\end{equation}
where $\mathrm{Quant}(\lVert\Delta\mathbf{a}_t\rVert)$ is an action-delta norm proxy, and $\mathrm{Hash}(\bar{\mathbf{v}}_t)$ is a lightweight vision hash for state matching. The robust hash normalizes $\bar{\mathbf{v}}_t$, applies a fixed random projection, and quantizes before hashing.

We distinguish a reuse request from an actual cache hit $h_t\in\{0,1\}$. When a hit occurs, we directly reuse the cached multimodal representation $\mathbf{z}_t$ and skip the VLM backbone forward, otherwise we compute $\mathbf{z}_t=f_{\mathrm{VLM}}(x_t,u)$ as usual. To keep cache population consistent with the router's intent, we only write back the newly computed $\mathbf{z}_t$ when reuse was requested but the lookup missed. 

\vspace{1mm}
\noindent\textbf{Token Pruning.}
\label{sec:method_compact}
Token gating determines which vision tokens should be retained. To obtain real wall-clock speedups beyond attention masking, we perform token pruning by physically removing pruned tokens and shortening the transformer sequence. Let $\mathbf{m}_t\in\{0,1\}^{N_v}$ denote the keep mask. Compaction produces
\begin{equation}
\begin{aligned}
(\tilde{\mathbf{V}}_t,\ \boldsymbol{\pi}_t)
&=\mathrm{Compact}(\mathbf{V}_t,\mathbf{m}_t),\\
\tilde{\mathbf{V}}_t &\in \mathbb{R}^{N'_v\times d_v},
\end{aligned}
\label{eq:compact}
\end{equation}
where $\boldsymbol{\pi}_t$ maps each kept token to its original patch index. We always keep at least one token to prevent degenerate empty sequences.

For Rotary Position Embedding (RoPE)-based backbones, naive reindexing after compaction would distort the original patch coordinates. We therefore preserve RoPE-consistent patch positions using $\boldsymbol{\pi}_t$:
\begin{equation}
\mathrm{pos}^{\mathrm{patch}}_{t,j}=1+\pi_{t,j},
\label{eq:rope_patch}
\end{equation}
and assign text positions to start after the original patch span $N_v^{\mathrm{orig}}$ recorded during compaction:
\begin{equation}
\mathrm{pos}^{\mathrm{text}}_{t,n}=1+N_v^{\mathrm{orig}}+n,
\quad n=0,\dots,T-2.
\label{eq:rope_text}
\end{equation}

During training, to stabilize optimization and maintain differentiability, we adopt a soft relaxation by scaling projected patch embeddings with token keep scores:
\begin{equation}
\mathbf{e}'_{t,i}=p^{\mathrm{topk}}_{t,i}\,\mathbf{e}_{t,i}.
\label{eq:soft_token_scale}
\end{equation}
Hard compaction is applied only at inference time for maximum speedup.

\vspace{1mm}
\noindent\textbf{Layer Skipping.}
\label{sec:method_skip}
We implement conditional depth execution by wrapping each transformer block with a lightweight gating mechanism. Given the hidden state $\mathbf{h}^{(\ell)}$ at layer $\ell$, the gated residual update is defined as
\begin{equation}
\mathbf{h}^{(\ell+1)}=
\mathbf{h}^{(\ell)}+\alpha_{t,\ell}(F_\ell(\mathbf{h}^{(\ell)})-\mathbf{h}^{(\ell)}),
\label{eq:layer_interp}
\end{equation}
where $F_\ell(\cdot)$ denotes the $\ell$-th transformer block and $\alpha_{t,\ell}\in[0,1]$ is the layer gate derived from $p^{\mathrm{lay}}_{t,\ell}$. During training, $\alpha_{t,\ell}$ remains soft; at inference, it is binarized so that inactive samples bypass the layer entirely.

For efficient execution, active samples are dynamically grouped into a sub-batch to run $F_\ell(\cdot)$, and the results are scattered back to the full batch.

\begin{algorithm}[t]
\caption{AC$^2$-VLA inference for a single control step $t$}
\label{alg:ap_infer}
\textbf{Input}: visual observation $x_t$, instruction $u$, previous action $\mathbf{a}_{t-1}$, step index $\tau_t$, cache $\mathcal{C}$\\
\textbf{Output}: action chunk $\mathbf{a}_{t:t+H}$
\begin{algorithmic}[1]
\STATE $\mathbf{V}_t \leftarrow f_{\mathrm{vis}}(x_t)$;\ $\bar{\mathbf v}_t \leftarrow \mathrm{MeanPool}(\mathbf{V}_t)$
\STATE $\mathbf{s}^v_t \leftarrow \tfrac{1}{2}(\mathrm{MeanPool}(\mathbf{V}_t)+\mathrm{MaxPool}(\mathbf{V}_t))$
\STATE $\mathbf{s}^u_t \leftarrow \mathrm{EmbedPool}(u)$
\STATE $\mathbf{c}_t \leftarrow f_{\mathrm{fuse}}(\psi_a(\mathbf{a}_{t-1}),\psi_v(\mathbf{s}^v_t),\psi_u(\mathbf{s}^u_t),\psi_\tau(\mathbf{e}(\tau_t)),\psi_c(\mathbf{s}^c_t))$\vspace{-3.3mm}
\STATE $(p^{\mathrm{cache}}_t,\mathbf{p}^{\mathrm{topk}}_t,\mathbf{p}^{\mathrm{lay}}_t) \leftarrow \mathcal{R}(\mathbf{c}_t)$
\STATE $h_t \leftarrow 0$
\IF{\textsc{ReuseReq}$(p^{\mathrm{cache}}_t)$}
    \STATE $\Delta\mathbf a_t \leftarrow \mathrm{DeltaProxy}(\mathbf{a}_{t-1})$
    \STATE $k_t \leftarrow (\mathrm{Quant}(\|\Delta\mathbf a_t\|),\,\mathrm{Hash}(\bar{\mathbf v}_t))$
    \STATE $(h_t,\mathbf{z}_t) \leftarrow \mathcal{C}.\mathrm{Get}(k_t)$
\ENDIF
\IF{$h_t=0$}
    \STATE $\mathbf{m}_t \leftarrow \mathrm{TopKMask}(\mathbf{p}^{\mathrm{topk}}_t)$;\ $\mathbf{g}_t \leftarrow \mathrm{BinGate}(\mathbf{p}^{\mathrm{lay}}_t)$
    \STATE $(\tilde{\mathbf V}_t,\boldsymbol{\pi}_t,N_v^{\mathrm{orig}}) \leftarrow \mathrm{Compact}(\mathbf V_t,\mathbf m_t)$
    \STATE $\mathrm{pos}_t \leftarrow \mathrm{RoPEAlign}(\boldsymbol{\pi}_t,N_v^{\mathrm{orig}})$
    \STATE $\mathbf{z}_t \leftarrow f_{\mathrm{VLM}}(x_t,u;\tilde{\mathbf V}_t,\mathrm{pos}_t,\mathbf g_t)$
    \IF{\textsc{ReuseReq}$(p^{\mathrm{cache}}_t)$}
        \STATE $\mathcal{C}.\mathrm{Put}(k_t,\mathbf{z}_t)$ \COMMENT{write back only on request \& miss}
    \ENDIF
\ENDIF
\STATE $\mathbf{a}_{t:t+H} \leftarrow p_\phi(\mathbf{a}\mid \mathbf{z}_t;\tau_t)$
\STATE \textbf{return} $\mathbf{a}_{t:t+H}$
\end{algorithmic}
\end{algorithm}

\begin{table*}[t]
\centering
\footnotesize
\setlength{\tabcolsep}{15pt}
\begin{threeparttable}
\begin{tabularx}{\textwidth}{c c c c c c c}
\toprule
\multirow{2}{*}{\textbf{Google Robot}} & \multirow{2}{*}{\textbf{Method}} & \multicolumn{5}{c}{\textbf{Success Rate ($\uparrow$)}} \\
\cmidrule(lr){3-7}
 & & \textbf{PickCan} & \textbf{MoveNear} & \textbf{Drawer} & \textbf{DrawerApple} & \textbf{Average} \\
\midrule
\multirow{7}{*}{\makecell{Visual Matching}}
  & RT-1        & 85.7\% & 44.2\% & 73.0\% & 6.5\%  & 52.4\% \\
  & RT-1-X      & 56.7\% & 31.7\% & 59.7\% & 21.3\% & 42.4\% \\
  & RT-2-X      & 78.7\% & 77.9\% & 25.0\% & 3.7\%  & 46.3\% \\
  & Octo-Base   & 17.0\% & 4.2\%  & 22.7\% & 0.0\%  & 11.0\% \\
  & OpenVLA     & 18.0\% & 56.3\% & 63.0\% & 0.0\%  & 34.3\% \\
  & CogACT      & 91.3\% & 85.0\% & 71.8\% & 50.9\% & 74.8\% \\
  & \textbf{AC$^2$-VLA}  & \textbf{97.2\%} & \textbf{82.7\%} & \textbf{80.6\%} & \textbf{46.8\%} & \textbf{76.8\%} \\
\midrule
\multirow{7}{*}{\makecell{Variant Aggregation}}
  & RT-1        & 89.8\% & 50.0\% & 32.3\% & 2.6\%  & 43.7\% \\
  & RT-1-X      & 49.0\% & 32.3\% & 29.4\% & 10.1\% & 30.2\% \\
  & RT-2-X      & 82.3\% & 79.2\% & 35.3\% & 20.6\% & 54.4\% \\
  & Octo-Base   & 0.6\%  & 3.1\%  & 1.1\%  & 0.0\%  & 1.2\% \\
  & OpenVLA     & 60.8\% & 67.7\% & 28.8\% & 0.0\%  & 39.3\% \\
  & CogACT      & 89.6\% & 80.8\% & 28.3\% & 46.6\% & 61.3\% \\
  & \textbf{AC$^2$-VLA}  & \textbf{88.7\%} & \textbf{84.4\%} & \textbf{28.2\%} & \textbf{45.1\%} & \textbf{61.6\%} \\
\bottomrule
\end{tabularx}
\end{threeparttable}
\caption{Google Robot success rates on SIMPLER under two evaluation settings. Most baseline results are reported by CogACT, and we add the AC$^2$-VLA row by evaluating our method under the same protocol.}
\label{tab:main_results}
\end{table*}

\subsection{Optimization}
\label{sec:method_opt}
We train the router to preserve dense-policy behavior under sparse execution with:
\begin{equation}
\mathcal{L}=\mathcal{L}_{distill}+\mathcal{L}_{reg}+\mathcal{L}_{temp}.
\label{eq:loss_total}
\end{equation}

\vspace{1mm}
\noindent\textbf{Action-guided self-distillation.}
We use a teacher-student scheme, where the teacher runs the dense policy and the student executes routed sparse inference, including cache reuse, token pruning, and layer skipping. We distill both action outputs and cognition features:
\begin{equation}
\mathcal{L}_{distill}=
\lambda_{\epsilon}\big\lVert \hat{\boldsymbol\epsilon}^{stu}-\hat{\boldsymbol\epsilon}^{tea}\big\rVert_2^2
+\lambda_{z}\mathcal{D}\!\left(\mathbf{z}^{stu}_t,\mathbf{z}^{tea}_t\right).
\label{eq:loss_distill}
\end{equation}
where $\hat{\boldsymbol\epsilon}$ denotes the action prediction, and $\mathbf{z}_t$ is the backbone representation. $\lambda_{\epsilon}$ and $\lambda_z$ control the relative weights, and $\mathcal{D}(\cdot,\cdot)$ is a feature-matching distance.

\vspace{1mm}
\noindent\textbf{Regularization and temporal smoothing.}
We add $\mathcal{L}_{reg}$ to enforce target token/layer budgets and supervise the reuse gate, and $\mathcal{L}_{temp}$ to penalize abrupt changes in gating decisions across timesteps for stable closed-loop control.

\begin{table*}[t]
\centering
\footnotesize
\setlength{\tabcolsep}{6.4pt}
\begin{threeparttable}
\begin{tabularx}{\textwidth}{c c c c c c c}
\toprule
\multirow{2}{*}{\textbf{WidowX Robot}} & \multirow{2}{*}{\textbf{Method}} & \multicolumn{5}{c}{\textbf{Success Rate ($\uparrow$)}} \\
\cmidrule(lr){3-7}
 & & \textbf{Put Spoon on Towel} & \textbf{Put Carrot on Plate} & \textbf{Stack Cube} & \textbf{Put Eggplant in Basket} & \textbf{Average} \\
\midrule
\multirow{6}{*}{\makecell{SIMPLER\\Visual Matching}}
  & RT-1-X     & 0.0\%  & 4.2\%  & 0.0\%  & 0.0\%  & 1.1\% \\
  & Octo-Base  & 15.8\% & 12.5\% & 0.0\%  & 41.7\% & 17.5\% \\
  & Octo-Small & 41.7\% & 8.2\%  & 0.0\%  & 56.7\% & 26.7\% \\
  & OpenVLA    & 4.2\%  & 0.0\%  & 0.0\%  & 12.5\% & 4.2\% \\
  & CogACT     & 71.7\% & 50.8\% & 15.0\% & 67.5\% & 51.3\% \\
  & \textbf{AC$^2$-VLA} & \textbf{71.2\%} & \textbf{58.0\%} & \textbf{14.8\%} & \textbf{74.0\%} & \textbf{54.5\%} \\
\bottomrule
\end{tabularx}
\end{threeparttable}
\caption{WidowX Visual Matching success rates on SIMPLER. Most baseline results are reported by CogACT, and we add the AC$^2$-VLA row by evaluating our method under the same protocol.}
\label{tab:widowx_results}
\end{table*}

\section{Experiment}

\subsection{Experimental Setup}

\vspace{1mm}
\noindent\textbf{Backbones.}
We build AC$^2$-VLA on CogACT \cite{CogACT}, a diffusion-based VLA model with a Prismatic-7B vision-language backbone and a DiT-Base action head.
To isolate routing effects, we freeze the pre-trained vision and language backbones and optimize only the lightweight routing modules, while keeping the action head unchanged with 8 denoising steps.

\vspace{1mm}
\noindent\textbf{Implementation Details.}
All experiments are conducted on a node with NVIDIA RTX 5090 GPUs.
We initialize from the CogACT-Base checkpoint \cite{CogACT} and train on the Bridge subset of Open X-Embodiment \cite{OpenX} for 3{,}000 steps using AdamW with batch size 48 and learning rate $1\times10^{-6}$.
We use an action horizon of $H=15$ with 8 diffusion steps.
Unless noted otherwise, AC$^2$-VLA enables action distillation, sets the maximum token pruning ratio to 0.6, and uses a cache reuse threshold of 0.2.

\vspace{1mm}
\noindent\textbf{Benchmarks.}
We evaluate on SIMPLER \cite{SIMPLER}, a high-fidelity simulation benchmark for robotic manipulation that aims to narrow the real-to-sim gap.
We report results on two robot embodiments under three protocols:
\begin{itemize}
  \item \textbf{Google Robot Visual Matching:} Tests generalization in visually matched real-world conditions on tasks including Pick Coke Can, Move Near, Open/Close Drawer, and Place Apple.
  \item \textbf{Google Robot Variant Aggregation:} Introduces variations in background, lighting, and distractors, providing a more challenging robustness setting.
  \item \textbf{WidowX Visual Matching:} Evaluates fine-grained manipulation on WidowX with tasks including Put Spoon on Towel, Put Carrot on Plate, Stack Cube, and Put Eggplant in Basket.
\end{itemize}
Following standard SIMPLER protocols, we use 3 Hz control with 513 Hz simulation for Google Robot and 5 Hz control with 500 Hz simulation for WidowX.
Episodes are capped at 80 steps for Google Robot and 120 steps for WidowX to penalize inefficient or stalled behaviors.

\begin{table*}[t]
  \centering
  \footnotesize
  \setlength{\tabcolsep}{8pt}
  \begin{tabular*}{\textwidth}{c c c c c c c c c}
    \toprule
    \multirow{2}{*}{\textbf{Setting}} & \multirow{2}{*}{\textbf{Method}} & \multicolumn{5}{c}{\textbf{Success Rate ($\uparrow$)}} & \multirow{2}{*}{\textbf{Speed-up ($\uparrow$)}} & \multirow{2}{*}{\textbf{FLOPs ($\downarrow$)}} \\
    \cmidrule(lr){3-7}
    & & \textbf{PickCan} & \textbf{MoveNear} & \textbf{Drawer} & \textbf{DrawerApple} & \textbf{Average} & & \\
    \midrule
    \multirow{6}{*}{\makecell{Visual\\Matching}}
      & CogACT       & 91.3\% & 85.0\% & 71.8\% & 50.9\% & 74.8\% & 1.00$\times$ & 100.0\% \\
      & VLA-Cache    & 92.0\% & 83.3\% & 70.5\% & 51.6\% & 74.4\% & 1.36$\times$ & 80.1\% \\
      & EfficientVLA & 95.3\% & 83.3\% & 70.3\% & 56.5\% & 76.4\% & 1.59$\times$ & 45.1\% \\
      & FastV        & 92.6\% & 81.4\% & 69.8\% & 52.4\% & 74.1\% & 1.21$\times$ & 42.0\% \\
      & MoLe-VLA     & 86.4\% & 80.2\% & 70.6\% & 50.4\% & 71.9\% & 1.53$\times$ & 47.4\% \\
      & \textbf{AC$^2$-VLA}   & \textbf{97.2\%} & \textbf{82.7\%} & \textbf{80.6\%} & \textbf{46.8\%} & \textbf{76.8\%} & \textbf{1.79$\times$} & \textbf{29.4\%} \\
    \midrule
    \multirow{6}{*}{\makecell{Variant\\Aggregation}}
      & CogACT       & 89.6\% & 80.8\% & 28.3\% & 46.6\% & 61.3\% & 1.00$\times$ & 100.0\% \\
      & VLA-Cache    & 91.7\% & 79.3\% & 32.5\% & 45.8\% & 62.3\% & 1.37$\times$ & 82.6\% \\
      & EfficientVLA & 94.8\% & 77.6\% & 28.4\% & 51.9\% & 63.2\% & 1.57$\times$ & 45.1\% \\
      & FastV        & 91.4\% & 78.6\% & 27.6\% & 50.6\% & 62.1\% & 1.19$\times$ & 42.0\% \\
      & MoLe-VLA     & 89.2\% & 79.5\% & 29.9\% & 46.2\% & 61.2\% & 1.49$\times$ & 46.3\% \\
      & \textbf{AC$^2$-VLA}   & \textbf{88.7\%} & \textbf{84.4\%} & \textbf{28.2\%} & \textbf{45.1\%} & \textbf{61.6\%} & \textbf{1.67$\times$} & \textbf{34.7\%} \\
    \bottomrule
  \end{tabular*}
  \caption{Comparison with efficiency-oriented VLA methods on SIMPLER across two settings.}
  \label{tab:efficiency_comparison}
\end{table*}

\vspace{1mm}
\noindent\textbf{Baselines.}
We compare AC$^2$-VLA with two groups of baselines: generalist dense VLA policies and efficiency-oriented methods.

\textbf{Generalist VLA Models:}
We report results for RT-1 \cite{RT1}, RT-2-X \cite{RT2}, Octo \cite{Octo}, OpenVLA \cite{OpenVLA}, and our backbone CogACT \cite{CogACT} in full precision, which serve as dense upper bounds on task success.

\textbf{Efficiency-Oriented Methods:}
We include representative acceleration approaches, including VLA-Cache \cite{VLACache} for temporal reuse, EfficientVLA \cite{EfficientVLA} for static pruning, MoLe-VLA \cite{MoLeVLA} for conditional layer skipping via mixture-of-layers routing, and FastV \cite{FastV} as a lightweight pruning baseline.
These comparisons characterize the trade-off between compute efficiency, measured by FLOPs and latency, and manipulation success across diverse tasks.

\subsection{Comparison with State-of-the-Art}
We compare AC$^2$-VLA on SIMPLER, reporting task success in Tables~\ref{tab:main_results} and~\ref{tab:widowx_results} and the speed--accuracy trade-off in Table~\ref{tab:efficiency_comparison}.

\vspace{1mm}
\noindent\textbf{Task Performance.}
AC$^2$-VLA achieves strong control performance across evaluation protocols. On Google Robot Visual Matching, it reaches 76.8\% average success, outperforming the dense CogACT baseline at 74.8\% and larger models such as RT-2-X at 46.3\%. Gains are most pronounced on precision-critical tasks, e.g., Drawer Opening improves from 71.8\% to 80.6\%, suggesting that action-prior-guided sparsification helps suppress distractors and stabilizes decision making. On Variant Aggregation, AC$^2$-VLA matches the full-precision baseline with 61.6\% versus 61.3\%, while consistently surpassing RT-1 and OpenVLA.

\vspace{1mm}
\noindent\textbf{Efficiency and Computational Cost.}
AC$^2$-VLA substantially reduces the inference cost of CogACT. As shown in Table~\ref{tab:efficiency_comparison}, it uses 29.4\% of the original FLOPs, yielding a 1.79$\times$ wall-clock speedup. Notably, this acceleration does not compromise performance and even improves success over dense CogACT, indicating that the removed computation largely corresponds to redundant or distracting features for closed-loop control.

\begin{figure}[t]
  \centering
  \includegraphics[width=0.95\columnwidth]{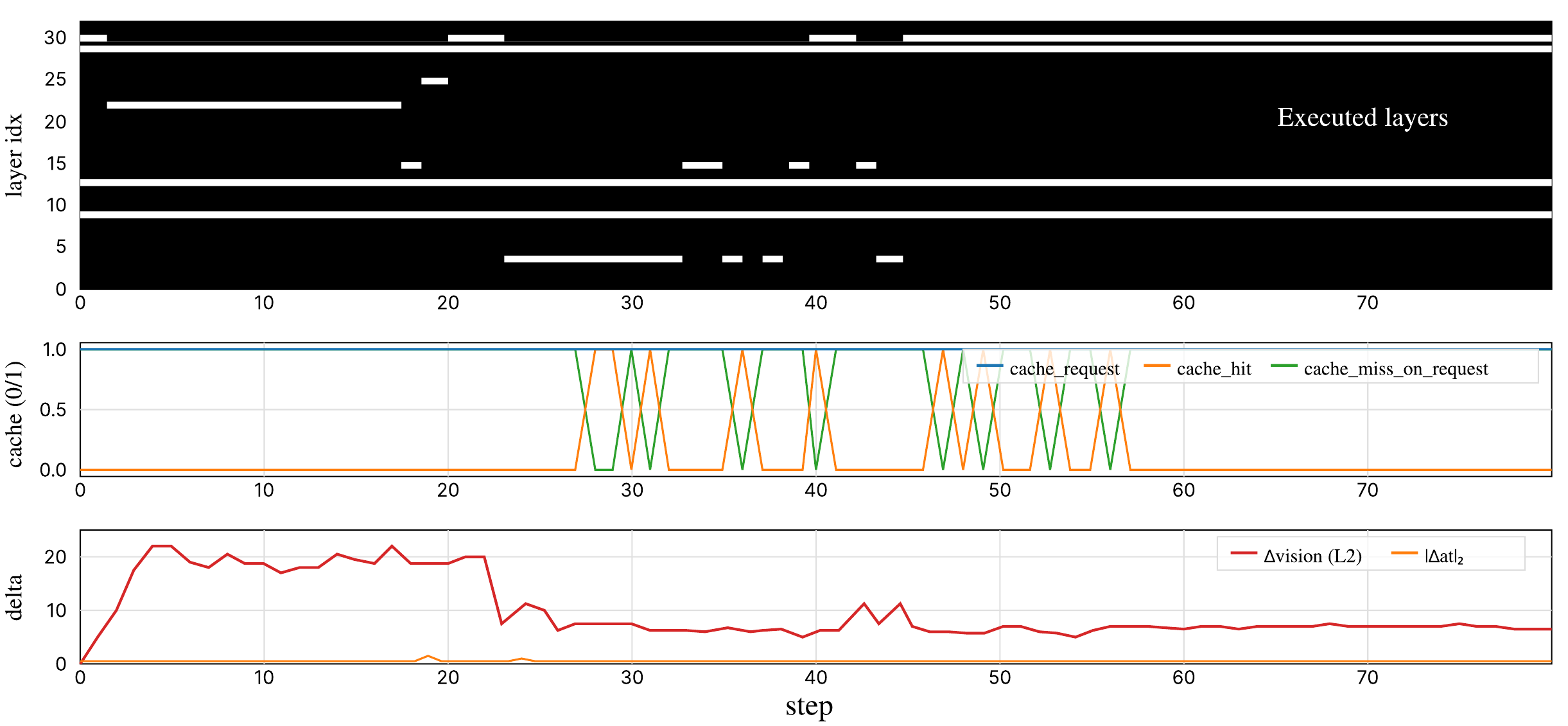}
  \vspace{-2mm}
  \caption{Adaptive layer execution and cache reuse over time.}
  \label{fig:layer_cache_over_time}
\end{figure}

\begin{figure}[t]
  \centering
  \begin{minipage}[t]{0.475\columnwidth}
    \centering
    \includegraphics[width=\linewidth]{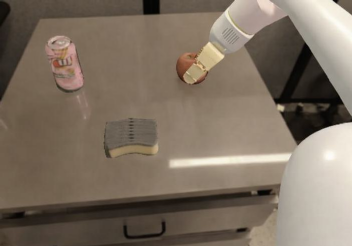}
  \end{minipage}\hfill
  \begin{minipage}[t]{0.475\columnwidth}
    \centering
    \includegraphics[width=\linewidth]{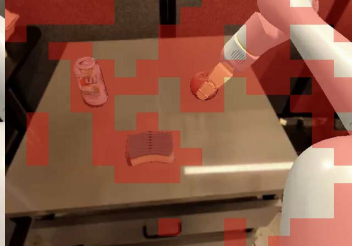}
  \end{minipage}
  \caption{Left: input observation. Right: visualization of token-level importance predicted by the action-conditioned router, highlighting regions relevant to the current manipulation stage while suppressing distractors. The highlighted regions adapt with the action context, focusing computation on interaction-critical areas.}
  \label{fig:token_importance_map}
\end{figure}

\begin{figure}[t]
  \centering
  \includegraphics[width=0.80\linewidth]{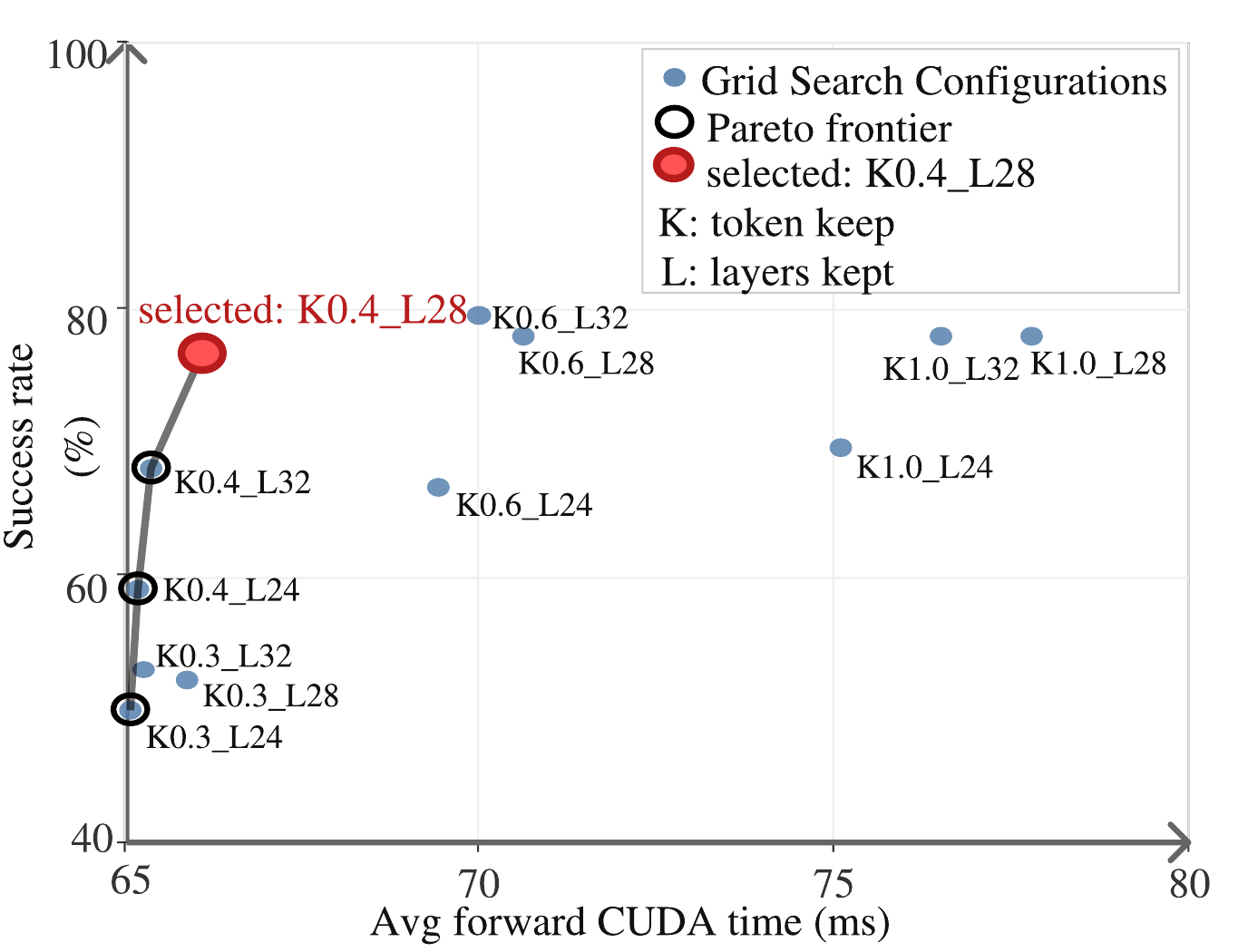}
  \caption{Pareto frontier for token pruning and layer skipping on the SIMPLER benchmark.}
  \label{fig:pareto_frontier}
\end{figure}

\subsection{Ablation Study}
\label{sec:ablation}
We ablate AC$^2$-VLA on SIMPLER Google Robot Visual Matching to validate key design choices.
We focus on the three efficiency axes controlled by the router, namely token pruning, layer skipping, and cognition reuse, and analyze their interaction under comparable budgets.
Ablation results are summarized in Table~\ref{tab:ablation_component}. We observe that each component provides complementary benefits, and the full model achieves the best speed-accuracy trade-off when all three axes are jointly enabled:

\begin{itemize}
    \item \textbf{Cache reuse.} Without cognition reuse, success drops to 70.5\% and speedup decreases to 1.66$\times$, suggesting that temporal reuse improves both efficiency and closed-loop stability. Fig.~\ref{fig:layer_cache_over_time} illustrates adaptive layer execution and cache hit behavior over time.
    \item \textbf{Token pruning.} Removing token pruning reduces speedup to 1.52$\times$, showing that spatial sparsification contributes most to acceleration. Fig.~\ref{fig:token_importance_map} visualizes the token-level routing patterns.
    \item \textbf{Layer routing.} Disabling layer routing drops the success rate to 67.4\% under similar FLOPs, indicating that conditional depth execution helps retain high-level reasoning while reducing compute.
    \item \textbf{Full model.} AC$^2$-VLA attains 76.8\% success while delivering a 1.79$\times$ speedup, demonstrating the effectiveness of jointly leveraging spatial, depth-wise, and temporal redundancies.
\end{itemize}

\subsection{More Exploration}
\label{sec:exploration}
Beyond component ablations, we further analyze the hyperparameter space and emergent behaviors of AC$^2$-VLA, focusing on the joint sparsity trade-off and the effect of cognition caching on closed-loop stability.

\vspace{1mm}
\noindent\textbf{Token-layer sparsity and the efficiency-accuracy trade-off.}
We perform a grid search over the token keep ratio $r_{topk}$ and the executed layer count $N_{lay}$, as shown in Fig.~\ref{fig:pareto_frontier}.
The results exhibit a clear Pareto frontier, where $r_{topk}=0.4$ and $N_{lay}=28$ achieves the best trade-off, reaching 1.79$\times$ speedup with 76.8\% success.
This suggests that many visual tokens are dispensable, while sufficient depth remains important for reasoning over the retained tokens.

\vspace{1mm}
\noindent\textbf{When cache reuse improves robustness beyond speed.}
Interestingly, cache reuse can improve robustness in addition to reducing compute.
As shown in Table~\ref{tab:ablation_sensitivity}, setting the cache threshold to $\tau_{cache}=0.2$ yields 87.1\% success, outperforming the dense baseline by +12.3\%.
We attribute this gain to improved temporal consistency: standard per-frame inference can amplify high-frequency visual noise and induce action jitter, whereas reusing $\mathbf{z}_t$ when the action context is stable effectively smooths decision making and stabilizes control.

\vspace{1mm}
\noindent\textbf{Sensitivity to key hyper-parameters.} We next vary the token keep ratio and the maximum executed depth to examine how accuracy degrades under more aggressive sparsification.
\begin{itemize}
    \item \textbf{Token sparsity.} Performance remains stable down to $r_{topk}=0.4$, but collapses at $r_{topk}=0.2$ with 33.3\% success, indicating a minimum visual information requirement for manipulation.
    \item \textbf{Depth.} The policy remains competitive with $N_{lay}=28$ at 77.3\% success, while reducing below 24 layers causes a sharp drop, suggesting that sufficient depth is critical for complex tasks.
\end{itemize}

\begin{table}[t]
\centering
  \setlength{\tabcolsep}{2pt}
  \resizebox{\columnwidth}{!}{
  \begin{tabular}{l c c c}
    \toprule
    \textbf{Configuration} & \textbf{Success Rate ($\uparrow$)} & \textbf{Speed-up ($\uparrow$)} & \textbf{FLOPs ($\downarrow$)} \\
    \midrule
    Dense baseline & 74.8\% & 1.00$\times$ & 100.0\% \\
    Full AC$^2$-VLA & 76.8\% & 1.79$\times$ & 29.4\% \\
    \midrule
    No cache reuse, $\tau_{cache}=1.0$ & 70.5\% & 1.66$\times$ & 38.6\% \\
    Without layer routing & 67.4\% & 1.68$\times$ & 29.4\% \\
    Without token pruning & 72.7\% & 1.52$\times$ & 66.8\% \\
    \bottomrule
  \end{tabular}
  }
  \vspace{-1.8mm}
  \caption{Component ablation on SIMPLER Google Robot Visual Matching. Speed-up is measured relative to the dense baseline.}
  \label{tab:ablation_component}
\end{table}

\begin{table}[t]
  \centering
  \setlength{\tabcolsep}{2pt}
  \resizebox{\columnwidth}{!}{%
  \begin{tabular}{l c c c c}
    \toprule
    \textbf{Sweep} & \textbf{Value} & \textbf{Success Rate ($\uparrow$)} & \textbf{Speed-up ($\uparrow$)} & \textbf{FLOPs ($\downarrow$)} \\
    \midrule
    \multirow{6}{*}{Token keep ratio $r_{topk}$} & 0.2 & 33.3\% & 1.69$\times$ & 25.8\% \\
     & 0.3 & 56.1\% & 1.66$\times$ & 34.7\% \\
     & 0.4 & 68.9\% & 1.63$\times$ & 44.1\% \\
     & 0.6 & 78.8\% & 1.47$\times$ & 62.5\% \\
     & 0.8 & 81.1\% & 1.26$\times$ & 81.0\% \\
    \midrule
    \multirow{6}{*}{Kept layers $N_{lay}$} & 22 & 69.7\% & 1.51$\times$ & 68.8\% \\
     & 24 & 73.5\% & 1.46$\times$ & 75.0\% \\
     & 26 & 78.0\% & 1.41$\times$ & 81.2\% \\
     & 28 & 77.3\% & 1.37$\times$ & 87.5\% \\
     & 30 & 80.3\% & 1.33$\times$ & 93.8\% \\
    \midrule
    \multirow{6}{*}{Cache threshold $\tau_{cache}$} & 0.00 & 82.6\% & 1.53$\times$ & 64.8\% \\
     & 0.05 & 81.1\% & 1.52$\times$ & 65.8\% \\
     & 0.10 & 77.3\% & 1.54$\times$ & 63.4\% \\
     & 0.20 & 87.1\% & 1.53$\times$ & 63.4\% \\
     & 0.30 & 78.8\% & 1.51$\times$ & 66.3\% \\
     & 0.40 & 79.5\% & 1.52$\times$ & 64.9\% \\
    \bottomrule
  \end{tabular}
  }
  \vspace{-1.8mm}
  \caption{Sensitivity sweeps on SIMPLER Google Robot Visual Matching, varying one efficiency component at a time with the other two disabled.}
  \label{tab:ablation_sensitivity}
\end{table}

\section{Conclusion}
We present AC$^2$-VLA, an action-context-aware framework for efficient Vision-Language-Action inference. By introducing a unified router that allocates computation across spatial, depth, and temporal dimensions based on the robot’s manipulation state, AC$^2$-VLA addresses the limitations of efficiency methods driven solely by visual complexity and enables adaptive closed-loop control. Experiments on the SIMPLER benchmark demonstrate a superior efficiency–accuracy trade-off, achieving a 1.79$\times$ speedup and reducing FLOPs to 29.4\% of the dense baseline while improving task success. These results indicate that action-guided sparsification acts as both an efficiency mechanism and a regularizer, suppressing visual distractors and promoting temporal consistency. Overall, AC$^2$-VLA suggests that aligning computation with action is a more effective paradigm for embodied intelligence than static compression, and points toward adaptive inference as a key ingredient for scalable generalist robot policies.

\appendix

\bibliographystyle{named}
\bibliography{references}

@article{OpenX,
  title         = {Open X-Embodiment: Robotic Learning Datasets and {RT-X} Models},
  author        = {Abby O’Neill and Abdul Rehman and Abhinav Gupta and Abhiram Maddukuri and Abhishek Gupta and Abhishek Padalkar and Abraham Lee and Acorn Pooley and Agrim Gupta and Ajay Mandlekar and et al.},
  journal       = {arXiv preprint arXiv:2310.08864},
  year          = {2023},
  doi           = {10.48550/arXiv.2310.08864},
  url           = {https://arxiv.org/abs/2310.08864}
}

@article{pi0,
  title         = {{$\pi_0$}: A Vision-Language-Action Flow Model for General Robot Control},
  author        = {Kevin Black and Noah Brown and Danny Driess and Adnan Esmail and Michael Equi and Chelsea Finn and Niccolo Fusai and Lachy Groom and Karol Hausman and Brian Ichter and Szymon Jakubczak and Tim Jones and Liyiming Ke and Sergey Levine and Adrian Li-Bell and Mohith Mothukuri and Suraj Nair and Karl Pertsch and Lucy Xiaoyang Shi and James Tanner and Quan Vuong and Anna Walling and Haohuan Wang and Ury Zhilinsky},
  journal       = {arXiv preprint arXiv:2410.24164},
  year          = {2024},
  doi           = {10.48550/arXiv.2410.24164},
  url           = {https://arxiv.org/abs/2410.24164}
}

@inproceedings{pi05,
  title         = {{$\pi_{0.5}$}: a Vision-Language-Action Model with Open-World Generalization},
  author        = {Kevin Black and Noah Brown and James Darpinian and Karan Dhabalia and Danny Driess and Adnan Esmail and Michael Robert Equi and Chelsea Finn and Niccolo Fusai and Manuel Y. Galliker and Dibya Ghosh and Lachy Groom and Karol Hausman and Brian Ichter and Szymon Jakubczak and Tim Jones and Liyiming Ke and Devin LeBlanc and Sergey Levine and Adrian Li-Bell and Mohith Mothukuri and Suraj Nair and Karl Pertsch and Allen Z. Ren and Lucy Xiaoyang Shi and Laura Smith and Jost Tobias Springenberg and Kyle Stachowicz and James Tanner and Quan Vuong and Homer Walke and Anna Walling and Haohuan Wang and Lili Yu and Ury Zhilinsky},
  booktitle     = {Proceedings of The 9th Conference on Robot Learning},
  pages         = {17--40},
  year          = {2025},
  editor        = {Joseph Lim and Shuran Song and Hae-Won Park},
  volume        = {305},
  series        = {Proceedings of Machine Learning Research},
  month         = {27--30 Sep},
  publisher     = {PMLR},
  url           = {https://proceedings.mlr.press/v305/black25a.html}
}

@article{TinyVLA,
  title         = {{TinyVLA}: Towards Fast, Data-Efficient {Vision-Language-Action} Models for Robotic Manipulation},
  author        = {Junjie Wen and Yichen Zhu and Jinming Li and Minjie Zhu and Kun Wu and Zhiyuan Xu and Ning Liu and Ran Cheng and Chaomin Shen and Yaxin Peng and Feifei Feng and Jian Tang},
  journal       = {arXiv preprint arXiv:2409.12514},
  year          = {2024},
  doi           = {10.48550/arXiv.2409.12514},
  url           = {https://arxiv.org/abs/2409.12514}
}

@article{SmolVLA,
  title         = {{SmolVLA}: A Vision-Language-Action Model for Affordable and Efficient Robotics},
  author        = {Mustafa Shukor and Dana Aubakirova and Francesco Capuano and Pepijn Kooijmans and Steven Palma and Adil Zouitine and Michel Aractingi and Caroline Pascal and Martino Russi and Andres Marafioti and Simon Alibert and Matthieu Cord and Thomas Wolf and R{\'e}mi Cad{\`e}ne},
  journal       = {arXiv preprint arXiv:2506.01844},
  year          = {2025},
  doi           = {10.48550/arXiv.2506.01844},
  url           = {https://arxiv.org/abs/2506.01844}
}

@article{FLOWER,
  title         = {{FLOWER}: Democratizing Generalist Robot Policies with Efficient Vision-Language-Action Flow Policies},
  author        = {Moritz Reuss and Hongyi Zhou and Marcel R{\"u}hle and {\"O}mer Erdin{\c{c}} Ya{\u{g}}murlu and Fabian Otto and Rudolf Lioutikov},
  journal       = {arXiv preprint arXiv:2509.04996},
  year          = {2025},
  doi           = {10.48550/arXiv.2509.04996},
  url           = {https://arxiv.org/abs/2509.04996}
}

@article{CogVLA,
  title         = {{CogVLA}: Cognition-Aligned Vision-Language-Action Model via Instruction-Driven Routing \& Sparsification},
  author        = {Wei Li and Renshan Zhang and Rui Shao and Jie He and Liqiang Nie},
  journal       = {arXiv preprint arXiv:2508.21046},
  year          = {2025},
  doi           = {10.48550/arXiv.2508.21046},
  url           = {https://arxiv.org/abs/2508.21046}
}

@article{LightVLA,
  title         = {The Better You Learn, The Smarter You Prune: Towards Efficient Vision-language-action Models via Differentiable Token Pruning},
  author        = {Titong Jiang and Xuefeng Jiang and Yuan Ma and Xin Wen and Bailin Li and Kun Zhan and Peng Jia and Yahui Liu and Sheng Sun and Xianpeng Lang},
  journal       = {arXiv preprint arXiv:2509.12594},
  year          = {2025},
  doi           = {10.48550/arXiv.2509.12594},
  url           = {https://arxiv.org/abs/2509.12594}
}

@article{FlashVLA,
  title         = {Think Twice, Act Once: Token-Aware Compression and Action Reuse for Efficient Inference in {Vision-Language-Action} Models},
  author        = {Xudong Tan and Yaoxin Yang and Peng Ye and Jialin Zheng and Bizhe Bai and Xinyi Wang and Jia Hao and Tao Chen},
  journal       = {arXiv preprint arXiv:2505.21200},
  year          = {2025},
  doi           = {10.48550/arXiv.2505.21200},
  url           = {https://arxiv.org/abs/2505.21200}
}

@article{SPVLA,
  title         = {{SP-VLA}: A Joint Model Scheduling and Token Pruning Approach for {VLA} Model Acceleration},
  author        = {Ye Li and Yuan Meng and Zewen Sun and Kangye Ji and Chen Tang and Jiajun Fan and Xinzhu Ma and Shutao Xia and Zhi Wang and Wenwu Zhu},
  journal       = {arXiv preprint arXiv:2506.12723},
  year          = {2025},
  doi           = {10.48550/arXiv.2506.12723},
  url           = {https://arxiv.org/abs/2506.12723}
}

@article{VOTE,
  title         = {{VOTE}: Vision-Language-Action Optimization with Trajectory Ensemble Voting},
  author        = {Juyi Lin and Amir Taherin and Arash Akbari and Arman Akbari and Lei Lu and Guangyu Chen and Taskin Padir and Xiaomeng Yang and Weiwei Chen and Yiqian Li and Xue Lin and David Kaeli and Pu Zhao and Yanzhi Wang},
  journal       = {arXiv preprint arXiv:2507.05116},
  year          = {2025},
  doi           = {10.48550/arXiv.2507.05116},
  url           = {https://arxiv.org/abs/2507.05116}
}

@article{CogACT,
  title         = {CogACT: A Foundational Vision-Language-Action Model for Synergizing Cognition and Action in Robotic Manipulation},
  author        = {Qixiu Li and Yaobo Liang and Zeyu Wang and Lin Luo and Xi Chen and Mozheng Liao and Fangyun Wei and Yu Deng and Sicheng Xu and Yizhong Zhang and Xiaofan Wang and Bei Liu and Jianlong Fu and Jianmin Bao and Dong Chen and Yuanchun Shi and Jiaolong Yang and Baining Guo},
  journal       = {arXiv preprint arXiv:2411.19650},
  year          = {2024},
  doi           = {10.48550/arXiv.2411.19650},
  url           = {https://arxiv.org/abs/2411.19650}
}

@article{SIMPLER,
  title         = {Evaluating Real-World Robot Manipulation Policies in Simulation},
  author        = {Xuanlin Li and Kyle Hsu and Jiayuan Gu and Karl Pertsch and Oier Mees and Homer Rich Walke and Chuyuan Fu and Ishikaa Lunawat and Isabel Sieh and Sean Kirmani and Sergey Levine and Jiajun Wu and Chelsea Finn and Hao Su and Quan Vuong and Ted Xiao},
  journal       = {arXiv preprint arXiv:2405.05941},
  year          = {2024},
  doi           = {10.48550/arXiv.2405.05941},
  url           = {https://arxiv.org/abs/2405.05941}
}

@article{RT1,
  title         = {RT-1: Robotics Transformer for Real-World Control at Scale},
  author        = {Anthony Brohan and Noah Brown and Justice Carbajal and Yevgen Chebotar and Joseph Dabis and Chelsea Finn and Keerthana Gopalakrishnan and Karol Hausman and Alex Herzog and Jasmine Hsu and Julian Ibarz and Brian Ichter and Alex Irpan and Tomas Jackson and Sally Jesmonth and Nikhil J. Joshi and Ryan Julian and Dmitry Kalashnikov and Yuheng Kuang and Isabel Leal and Kuang-Huei Lee and Sergey Levine and Yao Lu and Utsav Malla and Deeksha Manjunath and Igor Mordatch and Ofir Nachum and Carolina Parada and Jodilyn Peralta and Emily Perez and Karl Pertsch and Jornell Quiambao and Kanishka Rao and Michael Ryoo and Grecia Salazar and Pannag Sanketi and Kevin Sayed and Jaspiar Singh and Sumedh Sontakke and Austin Stone and Clayton Tan and Huong Tran and Vincent Vanhoucke and Steve Vega and Quan Vuong and Fei Xia and Ted Xiao and Peng Xu and Sichun Xu and Tianhe Yu and Brianna Zitkovich},
  journal       = {arXiv preprint arXiv:2212.06817},
  year          = {2022},
  doi           = {10.48550/arXiv.2212.06817},
  url           = {https://arxiv.org/abs/2212.06817}
}

@article{RT2,
  title         = {RT-2: Vision-Language-Action Models Transfer Web Knowledge to Robotic Control},
  author        = {Anthony Brohan and Noah Brown and Justice Carbajal and Yevgen Chebotar and Xi Chen and Krzysztof Choromanski and Tianli Ding and Danny Driess and Avinava Dubey and Chelsea Finn and Pete Florence and Chuyuan Fu and Montse {Gonzalez Arenas} and Keerthana Gopalakrishnan and Kehang Han and Karol Hausman and Alexander Herzog and Jasmine Hsu and Brian Ichter and Alex Irpan and Nikhil Joshi and Ryan Julian and Dmitry Kalashnikov and Yuheng Kuang and Isabel Leal and Lisa Lee and Tsang-Wei Edward Lee and Sergey Levine and Yao Lu and Henryk Michalewski and Igor Mordatch and Karl Pertsch and Kanishka Rao and Krista Reymann and Michael Ryoo and Grecia Salazar and Pannag Sanketi and Pierre Sermanet and Jaspiar Singh and Anikait Singh and Radu Soricut and Huong Tran and Vincent Vanhoucke and Quan Vuong and Ayzaan Wahid and Stefan Welker and Paul Wohlhart and Jialin Wu and Fei Xia and Ted Xiao and Peng Xu and Sichun Xu and Tianhe Yu and Brianna Zitkovich},
  journal       = {arXiv preprint arXiv:2307.15818},
  year          = {2023},
  doi           = {10.48550/arXiv.2307.15818},
  url           = {https://arxiv.org/abs/2307.15818}
}

@article{Octo,
  title         = {Octo: An Open-Source Generalist Robot Policy},
  author        = {{Octo Model Team} and Dibya Ghosh and Homer Walke and Karl Pertsch and Kevin Black and Oier Mees and Sudeep Dasari and Joey Hejna and Tobias Kreiman and Charles Xu and Jianlan Luo and You Liang Tan and Lawrence Yunliang Chen and Pannag Sanketi and Quan Vuong and Ted Xiao and Dorsa Sadigh and Chelsea Finn and Sergey Levine},
  journal       = {arXiv preprint arXiv:2405.12213},
  year          = {2024},
  doi           = {10.48550/arXiv.2405.12213},
  url           = {https://arxiv.org/abs/2405.12213}
}

@article{OpenVLA,
  title         = {OpenVLA: An Open-Source Vision-Language-Action Model},
  author        = {Moo Jin Kim and Karl Pertsch and Siddharth Karamcheti and Ted Xiao and Ashwin Balakrishna and Suraj Nair and Rafael Rafailov and Ethan Foster and Grace Lam and Pannag Sanketi and Quan Vuong and Thomas Kollar and Benjamin Burchfiel and Russ Tedrake and Dorsa Sadigh and Sergey Levine and Percy Liang and Chelsea Finn},
  journal       = {arXiv preprint arXiv:2406.09246},
  year          = {2024},
  doi           = {10.48550/arXiv.2406.09246},
  url           = {https://arxiv.org/abs/2406.09246}
}

@article{VLACache,
  title         = {VLA-Cache: Efficient Vision-Language-Action Manipulation via Adaptive Token Caching},
  author        = {Siyu Xu and Yunke Wang and Chenghao Xia and Dihao Zhu and Tao Huang and Chang Xu},
  journal       = {arXiv preprint arXiv:2502.02175},
  year          = {2025},
  doi           = {10.48550/arXiv.2502.02175},
  url           = {https://arxiv.org/abs/2502.02175}
}

@article{EfficientVLA,
  title         = {EfficientVLA: Training-Free Acceleration and Compression for Vision-Language-Action Models},
  author        = {Yantai Yang and Yuhao Wang and Zichen Wen and Luo Zhongwei and Chang Zou and Zhipeng Zhang and Chuan Wen and Linfeng Zhang},
  journal       = {arXiv preprint arXiv:2506.10100},
  year          = {2025},
  doi           = {10.48550/arXiv.2506.10100},
  url           = {https://arxiv.org/abs/2506.10100}
}

@article{FastV,
  title         = {An Image is Worth 1/2 Tokens After Layer 2: Plug-and-Play Inference Acceleration for Large Vision-Language Models},
  author        = {Chen, Liang and Zhao, Haozhe and Liu, Tianyu and Bai, Shuai and Lin, Junyang and Zhou, Chang and Chang, Baobao},
  journal       = {arXiv preprint arXiv:2403.06764},
  year          = {2024},
  doi           = {10.48550/arXiv.2403.06764},
  url           = {https://arxiv.org/abs/2403.06764}
}

@article{MoLeVLA,
  title         = {MoLe-VLA: Dynamic Layer-skipping Vision Language Action Model via Mixture-of-Layers for Efficient Robot Manipulation},
  author        = {Rongyu Zhang and Menghang Dong and Yuan Zhang and Liang Heng and Xiaowei Chi and Gaole Dai and Li Du and Yuan Du and Shanghang Zhang},
  journal       = {arXiv preprint arXiv:2503.20384},
  year          = {2025},
  doi           = {10.48550/arXiv.2503.20384},
  url           = {https://arxiv.org/abs/2503.20384}
}

@article{SQAPVLA,
  title         = {SQAP-VLA: A Synergistic Quantization-Aware Pruning Framework for High-Performance Vision-Language-Action Models},
  author        = {Hengyu Fang and Yijiang Liu and Yuan Du and Li Du and Huanrui Yang},
  journal       = {arXiv preprint arXiv:2509.09090},
  year          = {2025},
  doi           = {10.48550/arXiv.2509.09090},
  url           = {https://arxiv.org/abs/2509.09090}
}

\end{document}